\documentclass{article}
\usepackage{ijcai25}

\usepackage{times}
\usepackage{soul}
\usepackage{url}
\usepackage[hidelinks]{hyperref}
\usepackage[utf8]{inputenc}
\usepackage[small]{caption}
\usepackage{graphicx}
\usepackage{amsmath}
\usepackage{amsthm}
\usepackage{booktabs}
\usepackage{algorithm}
\usepackage{algorithmic}
\usepackage{pifont}
\usepackage[switch]{lineno}

\title{CoderAgent: Simulating Student Behavior for Personalized Programming Learning with Large Language Models}

\author{
Yi Zhan$^1$
\and
Qi Liu$^{1,2}$\thanks{Corresponding Author} \and
Weibo Gao$^{1}$\and
Zheng Zhang$^ 1$\and
Tianfu Wang$^ 1$\and \\
Shuanghong Shen$^ 2$\and 
Junyu Lu$^ 2$\And
Zhenya Huang$^{1,2}$
\\
\affiliations
$^1$State Key Laboratory of Cognitive Intelligence, University of Science and Technology of China\\
$^2$Institute of Artificial Intelligence, Hefei Comprehensive National Science Center
\emails
\{zy0119, weibogao, zhangzheng\}@mail.ustc.edu.cn,\{qiliuql, huangzhy\}@ustc.edu.cn,\\
\{ljunyu, shshen\}@iai.ustc.edu.cn,
tianfuwang.cs@gmail.com,
}

\begin{document}

\maketitle

\begin{abstract}

Personalized programming tutoring, such as exercise recommendation, can enhance learners' efficiency, motivation, and outcomes, which is increasingly important in modern digital education. However, the lack of sufficient and high-quality programming data, combined with the mismatch between offline evaluation and real-world learning, hinders the practical deployment of such systems.
To address this challenge, many approaches attempt to simulate learner practice data, yet they often overlook the fine-grained, iterative nature of programming learning, resulting in a lack of interpretability and granularity. To fill this gap, we propose a LLM-based agent, CoderAgent, to simulate students’ programming processes in a fine-grained manner without relying on real data. Specifically, we equip each human learner with an intelligent agent, the core of which lies in capturing the cognitive states of the human programming practice process. Inspired by ACT-R, a cognitive architecture framework, we design the structure of CoderAgent to align with human cognitive architecture by focusing on the mastery of programming knowledge and the application of coding ability. Recognizing the inherent patterns in multi-layered cognitive reasoning, we introduce the Programming Tree of Thought (PTOT), which breaks down the process into four steps: why, how, where, and what. This approach enables a detailed analysis of iterative problem-solving strategies. Finally, experimental evaluations on real-world datasets demonstrate that CoderAgent provides interpretable insights into learning trajectories and achieves accurate simulations, paving the way for personalized programming education.
\end{abstract}


\section{Introduction}


In the digital age, programming has emerged as a critical skill, fueling the growth of online educational platforms like \textit{LeetCode.com}, which have democratized access to programming education for learners worldwide \cite{www2025genmentor}. By analyzing user programming data, these platforms deliver personalized tutoring services, such as tailored exercise recommendations~\cite{zhao2023simulating,gao2025denoising}, contextual programming hints~\cite{marwan2019evaluation}, and interactive experimental simulations~\cite{gao2023research}, all designed to optimize learning efficiency and enhance the overall educational experience.

\begin{figure}[t]
  \centering
  \includegraphics[width=0.43\textwidth]{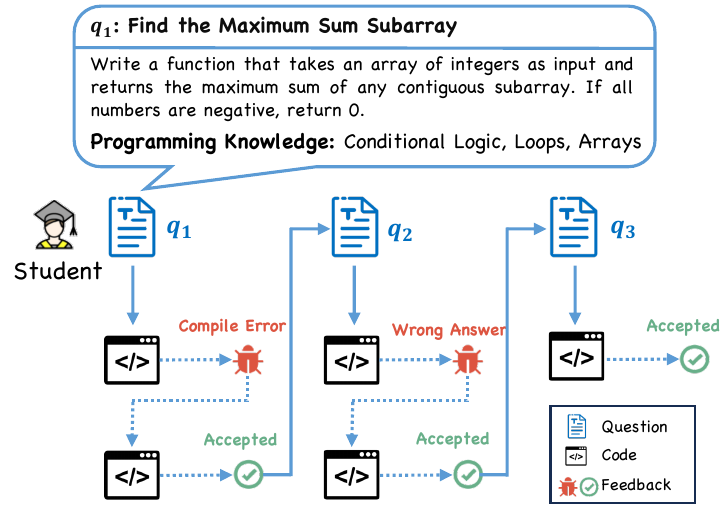}
  \caption{The example of the programming practice process}
  \label{fig:intro}
   \vspace{-0.2in}
\end{figure}

The effectiveness of these personalized tutoring services hinges on the availability of high-quality learner programming data, which is essential for training and evaluating adaptive algorithms. However, the inherent complexity and time-intensive nature of programming often result in slow progress for many learners, leading to a scarcity of such data. Compounding this issue, stringent privacy regulations further restrict the collection and utilization of learner data \cite{reddy2022ethical}, exacerbating the data shortage. This scarcity, coupled with the disconnect between historical offline-collected data and contemporary online learning practices, creates a significant disparity between offline evaluation metrics and real-world online performance. As a result, translating research insights into practical programming applications remains a formidable challenge.

To bridge this gap, simulating learner programming exercise data has emerged as a promising solution \cite{gao2025agent4edu}. Imagine an online programming platform equipped with a customizable coding simulation system that replicates how human learners iteratively modify their code when tackling programming challenges. Such a system could generate valuable simulated data to train and evaluate personalized tutoring algorithms. By analyzing these simulated interactions, algorithms could better adapt to diverse learner behaviors, narrowing the divide between simulated and real-world performance. Moreover, by anticipating potential code modifications, the system could offer more targeted guidance, such as improving code quality and fostering a more adaptive learning environment.

Existing methods for simulating learner exercises primarily focus on predicting whether a learner’s response to a given problem is correct \cite{li2022pst,puri2021codenet}. Still, they fall short of capturing the nuanced, iterative process of coding and revision. This limitation undermines their ability to provide meaningful insights for complex programming education. In reality, programming exercises involve a multifaceted learning process. As depicted in Figure~\ref{fig:intro}, learners write code to solve a specific problem (\textit{e.g., $q_1$}), receive system feedback (\textit{e.g.,compilation errors or correctness confirmation}), and iteratively refine their code based on this feedback. This cycle continues until the learner moves on to the next problem. Such complexity demands a more sophisticated simulator that goes beyond simplistic response prediction.
Furthermore, traditional simulators are predominantly data-driven, relying on large volumes of programming response data for training. This makes them ill-suited for cold-start scenarios, where data availability is limited, a common challenge in real-world applications. Recent advancements in large language models (LLMs) \cite{ijcai2024p236,yuan2024llms} have demonstrated their ability to emulate human-like intelligence. By leveraging LLMs to construct intelligent agents, it becomes possible to simulate complex human practice processes \cite{xu2024eduagent}. Additionally, the in-context learning capabilities of LLMs enable them to perform cold-start simulations with minimal reliance on real-world data \cite{huang2023recommender}, offering a promising solution to the limitations of current simulators. While some recent studies have begun exploring the use of LLMs to predict students' next code submissions, they still fail to capture the detailed, iterative nature of code modification. Consequently, there is a pressing need to develop an advanced agent capable of simulating the intricate, fine-grained process of programming problem-solving and code revision. 


In this paper, we leverage the intelligence of LLM agents to propose CoderAgent, a novel framework that can simulate students' programming processes in a fine-grained manner without relying on real data. Specifically, it includes Memory,  Tools, Planning \& Action, and Reflection modules. The \textbf{Memory} module is designed under the guidance of the ACT-R cognitive architecture \cite{anderson2014atomic,gao2021rcd}. This architecture posits that human programming behavior is primarily determined by programming-related cognitive factors and describes the cognitive structure of human programming as consisting of two components: the mastery of programming knowledge and the application of coding skills. As illustrated in Figure~\ref{fig:intro}, problem $q_1$ involves the knowledge concept conditional logic, loops and arrays, and thus the performance in solving $q_1$ is determined by the level of mastery of the relevant knowledge and coding ability. Based on this, we partition the memory into two regions that store programming knowledge and coding skills, respectively \cite{berges2012gap}, enabling CoderAgent to capture both the conceptual understanding and practical skills of learners. Additionally, by analyzing each student’s coding style and common errors from their past work, the framework generates code that mirrors the student’s realistic behavior. When a student encounters challenges beyond their current abilities, the agent simulates plausible mistakes they might make in subsequent iterations, providing insights into their learning process and informing the design of personalized programming exercises.
As for the \textbf{Planning \& Action} module, emerging from the inherent complexity of human learning \cite{Choi2009}, where learners must engage in multi-layered cognitive reasoning, from identifying problems to implementing solutions, we introduce the Programming Tree of Thought (PTOT), a novel reasoning mechanism specifically designed for programming tasks. PTOT decomposes debugging into four steps: \textit{why, how, where, and what}, pinpointing specific code segments for modification and leveraging the rich contextual information generated from \textbf{Tools} and programming process to enhance decision-making. By modeling the cognitive pathways students take when debugging, PTOT enables CoderAgent to achieve both high interpretability and fine-grained analysis of their behavior. Additionally, a Reflection module ensures the accuracy of the generated code. If the \textbf{Reflection} module detects that a modification exceeds the student’s capabilities or deviates from their coding profile, it prompts the agent to refine the output, ensuring alignment with the student’s abilities and coding style.

Finally, we evaluate CoderAgent using GPT-4o and Gpt-4o-mini APIs on several real-world programming datasets, demonstrating its ability to effectively simulate student coding behavior. We further apply CoderAgent to real-world programming tasks: mistake-prone point analysis and test case generation, showing its utility in practical coding applications. The performance of CoderAgent underscores its potential for improving programming education through personalized learning scenarios. Our main contributions are summarized as follows:
\begin{itemize}
    \vspace{-0.1cm}
    \item We introduce a novel LLM agent framework for programmer simulation that eliminates reliance on large-scale datasets, enabling effective performance even in data-scarce scenarios.
    \vspace{-0.1cm}
    \item We propose the Programming Tree of Thought (PTOT) to simulate realistic student code modifications, offering detailed insights into their learning process with interpretability and granularity.
    \vspace{-0.1cm}
    \item We validate our framework on real-world code submission data, showcasing its utility in personalized programming education.
\end{itemize}

\vspace{-0.0cm}
\section{Related Work}

\begin{figure*}[t]
  \centering
  \includegraphics[width=0.95\textwidth]{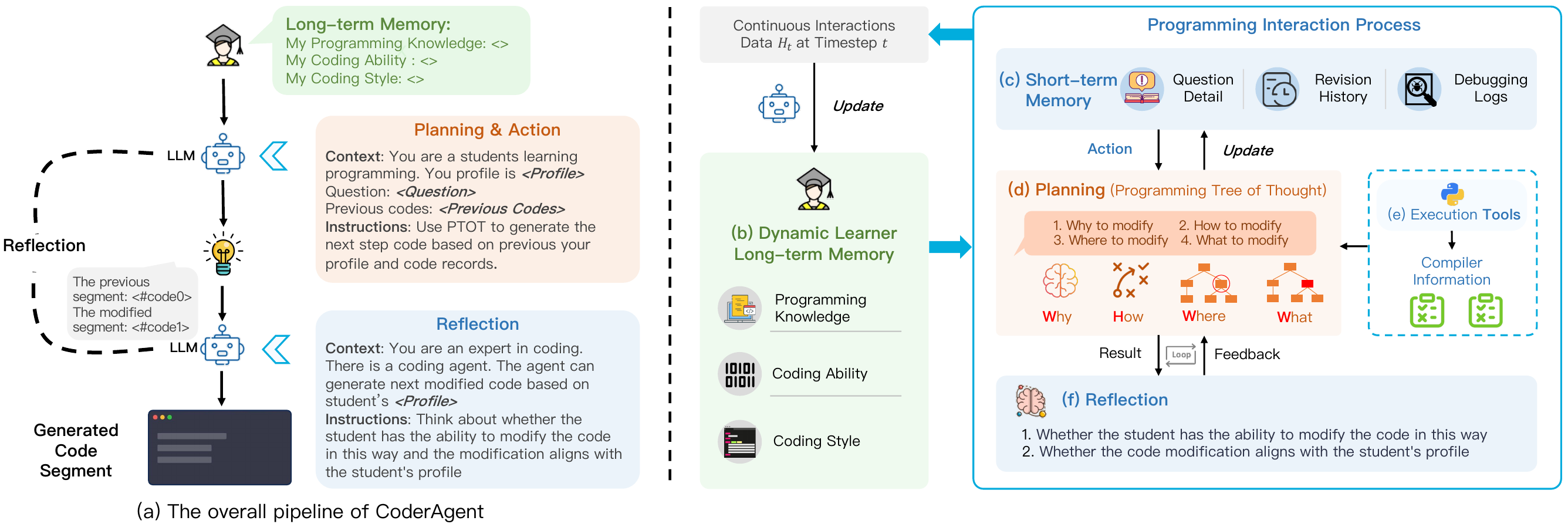}
  \vspace{-0.1cm}
  \caption{CoderAgent Framework. (a) The overall pipeline of CoderAgent; (b) Long-term memory, containing knowledge, ability and style; (c) Short-term memory, consisting of question, history and debugging logs; (d) Four steps of PTOT in planning module; (e) Execution tools, returning compiler information; (f) Reflection module, checking whether the output code is reasonable.}
  \label{fig:model}
  \vspace{-0.15cm}
\end{figure*}
\paragraph{Learner Response Simulation.}
Simulating human behavior has been a focal point in educational research \cite{ijcai2024p892,ijcai2024p996}, with early approaches relying on rule-based models and cognitive architectures, such as ACT-R \cite{anderson2014atomic}, to replicate students’ problem-solving processes. Knowledge Tracing (KT) \cite{corbett1994knowledge}  and Cognitive Diagnosis (CD) \cite{zhang2024understanding,zhangtowards} are foundational techniques in the field of educational data mining, focusing on modeling the evolution of a learner's knowledge state over time and simulating the final response of students. Recent studies in KT have explored more sophisticated models, such as deep learning-based approaches like Deep Knowledge Tracing \cite{piech2015deep,liu2019ekt,Liu2021}, which employ recurrent neural networks to capture complex temporal patterns in simulating behaviors. In the context of programming education, researchers have adapted KT frameworks to simulate students' mastery of programming concepts and skills \cite{wang2017learning,zhu2022programming,li2022pst,Liang2022HELPDKT}. A few of works, such as OKT \cite{liu2022open}, aim to predict students' code. Nonetheless, how to more precisely simulate how students modify their code remains an unresolved issue. In this study, we delve into capturing richer details about students' problem-solving strategies, misconceptions, and iterative refinement processes.

\vspace{-0.2cm}
\paragraph{LLM-based Agents.} Recently, LLMs have presented new opportunities to enhance human simulation \cite{aher2023using,park2023generative,zhao2023simulating,ijcai2024p1007}. The in-context learning capabilities of LLMs have been used as agents in various domains, including recommendation \cite{huang2023recommender} and education \cite{park2024empowering,qadir2023engineering,ijcai2024p627}. Edu4Agent \cite{gao2025agent4edu} represents the latest framework employing LLM agents to simulate student responses within the educational domain. However, its design targets general scenarios, making it unsuitable for simulating programming data. For programming education, although recent work \cite{liu2022open} has explored LLMs to predict students' code more specifically, their granularity is insufficient, and they heavily rely on data. Different from these approaches, to the best of our knowledge, we are the first to simulate the programming process through agents. By focusing on the incremental modifications students make to their code, we aim to replicate the cognitive processes underlying their problem-solving strategies. This simulation framework provides a more accurate representation of how students tackle programming tasks. Such insights are critical for developing adaptive educational tools and interventions tailored to individual learning trajectories.

\vspace{-0.2cm}
\section{Methodology}
In this section, we present the proposed framework, named CoderAgent. Our approach enables LLM agents to simulate the programming process. The overall framework of our proposed CoderAgent is depicted in Figure \ref{fig:model}.
\subsection{Preliminaries}
\paragraph{Problem Definition.}
In our study, we define the programming history of a student as a sequence of code submissions, denoted as a programming history sequence \( H_n = \{h_1, h_2, \dots, h_n\} \), where each \( h_i = (e_i, c_i, f_i) \) represents the code \( c_i \) submitted by the student at step \( i \) for a specific exercise \( e_i \). Here, \( c_i \) is the code submitted by the student, \( e_i \) refers to the exercise associated with that submission, and \( f_i \) is the correctness indicator of the submission. Specifically, \( f_i = 1 \) if the submission passes all tests and meets the requirements of the exercise, and \( f_i = 0 \) if the submission fails to meet the requirements. Each code submission represents an incremental change in the student's code, reflecting their attempts to solve the exercise and address previous errors. The primary goal of our study is to predict the student's next code submission as well as whether it will ultimately result in a correct solution. Formally, given the historical sequence of code submissions \( H_n = \{h_1, h_2, \dots, h_n\} \), our objective is to predict the student's next code submission \( c_{n+1} \) and determine whether it will ultimately be correct. While the intermediate submissions \( c_1, c_2, \dots, c_n \) are part of the student’s process, the focus is on predicting the student's next submission and determining if it leads to a correct final solution.

\paragraph{Overview of CoderAgent.} The CoderAgent framework consists of five key modules: Memory, Tools, Planning, Action, and Reflection, each designed to simulate the iterative and dynamic coding process. Rooted in the ACT-R theory from cognitive psychology, the Memory module categorizes a student’s programming proficiency into programming knowledge and coding skills, enabling the agent to store and recall code snippets, mistakes, and adjustments for informed decision-making. The Planning module incorporates the Programming Tree of Thought (PTOT), an innovative structure that captures the iterative progression of a student’s thought process during coding, offering fine-grained interpretability and aligning with cognitive reasoning to enhance realism. Tools provide compilers for various programming languages, delivering feedback on compilation errors, while the Action module governs step-by-step coding strategies, including writing, testing, and refining code. Finally, the Reflection module evaluates the agent’s output, ensuring alignment with the student’s abilities and refining problem-solving approaches through iterative feedback. Together, these modules create a comprehensive and personalized simulation of the student coding process.

\subsection{Memory Module}
The memory module is specifically designed to address the unique demands of code generation and iterative programming, setting it apart from memory systems in prior studies. Unlike traditional agents that operate on broad problem-solving tasks, our framework must manage the intricacies of programming, such as handling syntax, debugging, and iterative refinements. To meet these challenges, we divide memory into long-term memory and short-term memory, each tailored to the dynamics of the programming process.
\paragraph{Long-term Memory.} The long-term memory component is designed to simulate the foundational knowledge and skills that students develop throughout their programming journey. Grounded in the ACT-R theory from cognitive psychology, the acquisition of expertise is conceptualized as comprising two primary elements: knowledge and ability. Applied to programming, these correspond to programming knowledge (\textit{e.g., understanding syntax, structures, and algorithms}) and coding ability (\textit{e.g., implementing, debugging, and optimizing code}). Research \cite{mckeithen1981knowledge} has shown that students at different skill levels demonstrate distinct hierarchical structures in their programming knowledge. Expert-level students, for example, tend to organize programming constructs (such as keywords or syntax rules) into meaningful hierarchies, enabling more efficient recall and application. This hierarchical knowledge structure significantly enhances their coding ability, leading to more precise and effective programming. In our framework, the long-term memory module reflects this layered organization of programming knowledge and captures its influence on coding ability. Beyond these core factors, the module also records individual coding styles and common errors for each student. This personalized profiling ensures that the generated code closely aligns with the student’s natural tendencies, enabling a more realistic simulation of how students might iteratively modify code or make specific mistakes during programming.
By continuously updating the stored programming knowledge, coding ability, style, and common errors using historical data from students' prior coding tasks, we ensure that the memory module evolves to increasingly reflect the real-world behaviors of individual students. During training, the framework updates $M_{LT}^i$ iteratively using historical task data $H^i_t$ at time $t$:
\begin{align}
M_{LT}^i(t+1) = f(M_{LT}^i(t), H^i_t)
\end{align}
where $f$ is the update function that integrates new insights from task data $D^i_t$ to refine the student’s programming knowledge, ability, style, and error profiles.
\paragraph{Short-term Memory.}In contrast, short-term memory captures transient information related to the current task, such as recent question details, written code snippets and debugging history. This allows the agent to maintain a clear and updated context throughout the iterative coding process. For example, when resolving a compilation error, the short-term memory retains the feedback from the compiler and integrates it into subsequent edits. Similarly, it stores intermediate results during debugging or optimization, enabling the agent to refine the code systematically without losing sight of prior changes. The adaptability of short-term memory makes it essential for managing the back-and-forth nature of programming tasks. This dynamic capability is key to creating realistic simulations of student coding behavior.
\subsection{Programming Tools}
The tools module in CoderAgent is responsible for interfacing with external compilers to simulate the real-world feedback loop students rely on during programming. This module enables the agent to compile code snippets, interpret error messages, and retrieve translation information from various programming languages. By leveraging these tools, the agent gains access to detailed feedback that informs the iterative refinement process of the code. For example, when the agent submits code to a compiler, the Tools module processes the compilation results, including syntax errors, runtime issues, or warnings. These results are then fed back into the agent’s  planning  module, allowing it to adjust its strategy and improve the code accordingly. By simulating the practical utility of compilers, the tools module enhances the realism and functionality of the CoderAgent, enabling it to mimic how students utilize external feedback to iteratively refine their programming solutions.

\subsection{Elaborate Planning \& Action}
The planning \& action module is the decision-making core of the CoderAgent. It is designed to replicate the strategic and procedural steps students take during programming. This module orchestrates the agent's actions by breaking down complex programming tasks into manageable subtasks, selecting the appropriate steps, and executing them iteratively.
\paragraph{Planning.}
Inspired by prior research on Chain-of-Thought (CoT) \cite{wei2022chain} reasoning, we propose a specialized framework for the programming domain: \textbf{Programming Tree of Thought (PTOT)}. PTOT is designed to address the unique characteristics of the iterative nature of code development. Unlike other learning scenarios, the process in programming is often localized. When a student receives error feedback from a compiler, they typically only need to revise specific code snippets to resolve the issue. Even when facing semantic errors, adjustments are usually confined to small segments of the code rather than requiring wholesale rewrites. Regardless of whether these adjustments succeed or introduce new errors, this localized iteration forms the core of a student’s problem-solving approach in programming.
To replicate this iterative reasoning, PTOT breaks down the student’s thought process into four key steps:
\begin{enumerate}
    \item \textbf{Why do I need to modify the code?}  
    This step identifies the root cause of the issue based on feedback, such as errors flagged by the compiler. It involves understanding whether the error arises from syntax or semantics.
    
    \item \textbf{How do I modify the code?}  
    Here, the agent formulates a strategy for addressing the issue. For example, it decides whether to rewrite the logic, change variable declarations, or update function parameters.
    
    \item \textbf{Where should I modify the code?}  
    This step pinpoints the location that requires changes, guided by compiler messages or a structural understanding of the code.
    
    \item \textbf{What should I modify in the code?}  
    Finally, the agent determines the precise changes to be made, such as editing lines or correcting incorrect data types.
\end{enumerate}
To facilitate this process, the LLM agent integrates multiple sources of input: the latest version of the student’s code $c_t$, exercise or problem description $e_i$ and feedback from the Tools module $\mathcal{F}_t$. Using these inputs, the planning module formulates a structured roadmap that directs the subsequent action module. This roadmap ensures that each step is logically grounded and consistent with the feedback received. By modeling this iterative and localized thought process, PTOT enables CoderAgent to effectively simulate the nuanced way students approach problem-solving in tasks.

\paragraph{Action.}
The action module is responsible for executing the specific steps outlined in the Planning module. This includes identifying the most likely code segment requiring modification and generating its replacement. Guided by the planning module's structured roadmap and real-time feedback from the tools module, the action module ensures that the modifications align with the coding objective while addressing the highlighted issues. The action module identifies the code segment  $s_{t}$  within  $c_{t}$  that is most likely the source of the issue and proposes a replacement  $s_{t}^{\prime}$  based on planned modification strategy $p_{t}$. Formally, the output of the action module can be represented as:
\begin{equation}
    s_{t}^{\prime}=g\left(s_{t}, p_{t}\right),
\end{equation}
where $g$ is a transformation function that generates  $s_{t}^{\prime}$ by applying planned modifications $p_{t}$  to the identified code segment $ s_{t}$.
where $g$ applies planned modifications $p_t$ to code segment $s_t$ to generate $s_{t}^{\prime}$.
\subsection{Ability Reflection}
Despite the advanced reasoning capabilities of LLMs, they can occasionally exhibit errors in judgment or decision-making. For instance, a LLM agent might suggest a code modification that appears syntactically correct but is beyond the user's skill level or inconsistent with their typical coding style. To address such issues, prior research has incorporated self-reflection mechanisms into LLMs, enabling them to identify and mitigate errors autonomously. Inspired by this approach, we have developed a Reflection module to enhance the realism and accuracy of the CoderAgent framework. The execution steps of the reflection module are listed as follows:
\begin{itemize}
    \vspace{-0.1cm}
    \item \textbf{Whether the student has the ability to modify the code in this way.} This step assesses whether the proposed modification aligns with the student’s coding abilities, as captured in the memory module. If the modification requires advanced knowledge that exceed the student's profiled skill level, the reflection module flags the change as inappropriate. 
    
    \vspace{-0.1cm}
    \item \textbf{Whether the code modification aligns with the student's profile.} Beyond correctness, the reflection module evaluates whether the modification adheres to the student's typical patterns of problem-solving. By referencing long-term memory, the module ensures that the modifications remain consistent with their prior work.
    
\end{itemize}
Through this two-step evaluation, the reflection module identifies potential mismatches between the agent’s outputs and the student’s profile. When discrepancies are detected, the module provides feedback to the planning module, prompting the agent to revise its strategy. This iterative feedback loop reduces reasoning errors and improves the agent’s ability to emulate the student’s thought process accurately.

\begin{table}[t]
\caption{The statistics of the datasets.}
\vspace{-0.1in}
 \centering
 \resizebox{0.9\linewidth}{!}{%
\begin{tabular}{c|ccc}
\toprule
Dataset             & CodeNet     & CSEDM   \\ \midrule
\# Students             & 137,833     & 506       \\
\# Exercises             & 4,053     & 50      \\
\# Solutions           & 13,916,868   & 125,578    \\
Avg. exercise number of learner          & 3,437.96             & 47.49          \\  \bottomrule
\end{tabular}
}
 \label{static}
 \vspace{-0.2cm}
\end{table}
\begin{table}[t]
\caption{Tasks that can be accomplished by different methods.}
\vspace{-0.1in}
\centering
 \resizebox{0.8\linewidth}{!}{%
\begin{tabular}{ll|cccc}
\midrule
\multicolumn{2}{l|}{}           & Task 1 & Task 2 & Task 3 & Task 4 \\ \midrule
\multicolumn{2}{l|}{DKT}        & -      & -      & -      &    \ding{52}     \\ 
\multicolumn{2}{l|}{codeBERT}   & -      & -      & -      &    \ding{52}     \\ 
\multicolumn{2}{l|}{PDKT}       & -      & -      & -      &   \ding{52}      \\ 
\multicolumn{2}{l|}{PST}        & -      & -      & -      &    \ding{52}     \\ 
\multicolumn{2}{l|}{Agent4Edu}  & -      & -      & -      &    \ding{52}     \\ 
\multicolumn{2}{l|}{OKT}        & \ding{52}    &     \ding{52}    &  \ding{52}       & -      \\ 
\multicolumn{2}{l|}{CoderAgent} &  \ding{52}       &   \ding{52}      &    \ding{52}     &     \ding{52}     \\ \bottomrule
\end{tabular}
}
\label{table:result1}
\vspace{-0.35cm}
\end{table}
\section{Experiments}

\begin{table*}[t]
  \caption{Results of comparison methods on different task. Only OKT and CoderAgent can complete task 1, 2 and 3. The best results are bold, the second-best results are marked by an underline, and $\uparrow$ means the higher score the better performance.}
\vspace{-0.1in}
\centering
\resizebox{0.75\linewidth}{!}{
\begin{tabular}{ll|ccc|ccc}
\midrule
\multicolumn{2}{c|}{}                         & \multicolumn{3}{c|}{CSEDM}                                                                         & \multicolumn{3}{c}{CodeNet}                                                                       \\ \midrule
\multicolumn{2}{c|}{} & \multicolumn{1}{c|}{Task 1} & \multicolumn{1}{c|}{Task 2} & \multicolumn{1}{c|}{Task 3}    & \multicolumn{1}{c|}{Task 1} & \multicolumn{1}{c|}{Task 2} & \multicolumn{1}{c}{Task 3}    \\ \hline
\multicolumn{2}{c|}{}                         & \multicolumn{1}{c|}{ACC $\uparrow$}       & \multicolumn{1}{c|}{ACC $\uparrow$}       & \multicolumn{1}{c|}{CodeBLEU $\uparrow$}     & \multicolumn{1}{c|}{ACC $\uparrow$}       & \multicolumn{1}{c|}{ACC $\uparrow$}       & \multicolumn{1}{c}{CodeBLEU $\uparrow$}  \\ \midrule
\multicolumn{2}{l|}{OKT}                      & \multicolumn{1}{c|}{0.2531}       & \multicolumn{1}{c|}{0.4430}       & \multicolumn{1}{c|}{0.5115}        & \multicolumn{1}{c|}{0.2063}       & \multicolumn{1}{c|}{0.2222}       & \multicolumn{1}{c}{0.1805}         \\ \midrule
\multicolumn{2}{l|}{CoderAgent(4o-mini)}                      & \multicolumn{1}{c|}{\underline{0.3670}}       & \multicolumn{1}{c|}{\underline{0.5229}}       & \multicolumn{1}{c|}{\underline{0.7580}}     & \multicolumn{1}{c|}{\textbf{0.4464}}      & \multicolumn{1}{c|}{\textbf{0.5893}}       & \multicolumn{1}{c}{\textbf{0.5946}}    \\ \multicolumn{2}{l|}{CoderAgent(4o)}                      & \multicolumn{1}{c|}{\textbf{0.3841}}     & \multicolumn{1}{c|}{\textbf{0.5324}}      & \multicolumn{1}{c|}{\textbf{0.7698}}    & \multicolumn{1}{c|}{\underline{0.4010}}       & \multicolumn{1}{c|}{\underline{0.4853}}       & \multicolumn{1}{c}{\underline{0.5900}}   \\ \bottomrule
\end{tabular}
}
\label{table:result2}
\vspace{-0.3cm}
\end{table*}

\begin{table}[t]
\caption{The AUC scores of comparison methods on task 4.}
\vspace{-0.1in}
\centering
\begin{tabular}{ll|c|c}
\midrule
\multicolumn{2}{c|}{}           & CSEDM  & CodeNet \\
\midrule
\multicolumn{2}{l|}{DKT}        & 0.5184 & 0.5239  \\ 
\multicolumn{2}{l|}{codeBERT}   & 0.5142 & 0.5255  \\ 
\multicolumn{2}{l|}{PDKT}       & 0.5091 & 0.5078  \\ 
\multicolumn{2}{l|}{PST}        & 0.5261 & 0.5272  \\ 
\multicolumn{2}{l|}{Agent4Edu}  & 0.5071 & 0.5116  \\ \midrule
\multicolumn{2}{l|}{CoderAgent(4o-mini)}  & \underline{0.5351} & \underline{0.5386}  \\
\multicolumn{2}{l|}{CoderAgent(4o)} & \textbf{0.5493} & \textbf{0.5522}  \\ \bottomrule
\end{tabular}
\label{table:result3}
\vspace{-0.25cm}
\end{table}
\subsection{Experimental Setup}
\paragraph{Datasets.} Following prior research \cite{li2022pst,liu2022open}, we utilized two datasets for experiments: CodeNet \cite{puri2021codenet} and CSEDM. The basic data statistics are presented in Table \ref{static}. CodeNet, provided by IBM, comprises programming submissions sourced from two popular online coding platforms, AIZU.org and Atcoder.org. Given the diversity of programming languages in CodeNet, we restricted our analysis to the Python subset, ensuring a more focused and computationally efficient evaluation. CSEDM is a publicly available, college-level dataset containing real-world student programming submissions with Java language. To maintain data quality, we excluded students with an insufficient number of submissions, concentrating on those with richer activity records. Furthermore, we sampled subsets from both datasets for experiments due to the expensive cost of API calls associated with processing. Both datasets were split into 80\% training, 10\% validation, and 10\% testing.

\paragraph{Evaluation.}
We define four tasks for the evaluation:
\begin{itemize}
\vspace{-0.1cm}

    \item \textbf{Task 1: }Predict the next modification intention.
\vspace{-0.1cm}
    
    \item \textbf{Task 2: }Predict where will be modified next.
\vspace{-0.1cm}
    
    \item \textbf{Task 3: }Predict the student’s next code submission.
\vspace{-0.1cm}
    
    \item \textbf{Task 4: }Predict whether the next code submission will result in an AC (Accepted) outcome. 
\end{itemize}
 Task 1 and task 2 is utilized to evaluate the fine-grained precision of code iteration, assessing whether the system can accurately simulate what students think and do. We leverage LLMs (GPT-4o-mini) to assess the accuracy (ACC) of these two tasks. For task 3, we use metrics that evaluate the framework’s ability to generate code that closely matches the actual student submission in the test set. Specifically, we employ CodeBLEU \cite{ren2020codebleu}, an adaptation of the traditional BLEU metric tailored for code evaluation. CodeBLEU assesses the similarity between the predicted and actual code by considering both syntactic and semantic aspects, providing a comprehensive measure of prediction quality.  For task 4, the performance of the framework on this task is measured using Area Under the Curve (AUC), which evaluates the model's ability to distinguish between positive and negative outcomes across various thresholds.

\paragraph{Methods.} 
The details of the above baselines are as follows:
\begin{itemize}
    \item DKT \cite{piech2015deep} leverages RNN to assess knowledge mastery.
    \vspace{-0.05cm}
    \item codeBERT \cite{feng2020codebert} is a model for code representation learning. To track programming skill progression, we leverage codeBERT to represent the source code, and incorporate code embeddings, exercise embeddings, and feedback embeddings to model the complete learning sequence of students using a LSTM \cite{hochreiter1997long} structure.
    \vspace{-0.05cm}
    \item PDKT \cite{puri2021codenet} utilizes the exercise information and the source code to implement double-sequence modeling process to track programming knowledge. 
    \vspace{-0.05cm}
    \item PST \cite{li2022pst} divided programming skill into programming knowledge and coding ability to get more finegrained assessment.
    \vspace{-0.05cm}
    \item Agent4Edu \cite{gao2025agent4edu} is the only prior work to use LLM agents to simulate student's response. In our experiments, we did not use the cognitive diagnosis tools employed in the original paper, and we plan to use the full version of Agent4Edu in future research.
    \vspace{-0.05cm}
    \item OKT \cite{liu2022open} is the only framework which utilizes LLMs to track programming knowledge.
\end{itemize}
We use GPT-4o(2024-11-20) and GPT-4o-mini(2024-07-18) as the CoderAgent core. Our code and the simulated data is partially available at \url{https://github.com/USTChandsomeboy/CoderAgent}.
\begin{table*}[t]
\vspace{-0.1in}
\caption{The specific iterative process of CoderAgent modifying the code}
\vspace{-0.1in}
\centering
 \resizebox{1\linewidth}{!}{%
\begin{tabular}{|c|c|c|c|}
\hline
\textbf{Submisson 1} & \textbf{Submisson 2} & \textbf{Submisson 3} & \textbf{Submisson  4}  \\
\hline
\includegraphics[width=0.22\textwidth]{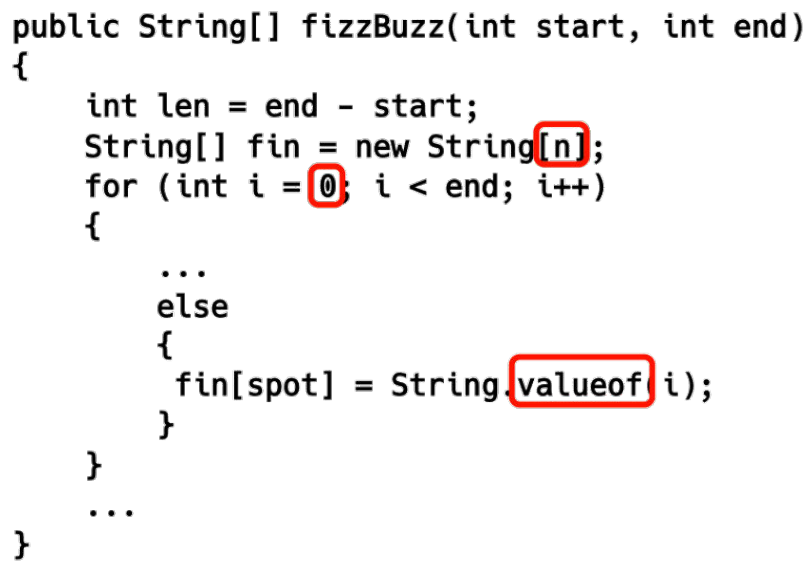} & 
\includegraphics[width=0.22\textwidth]{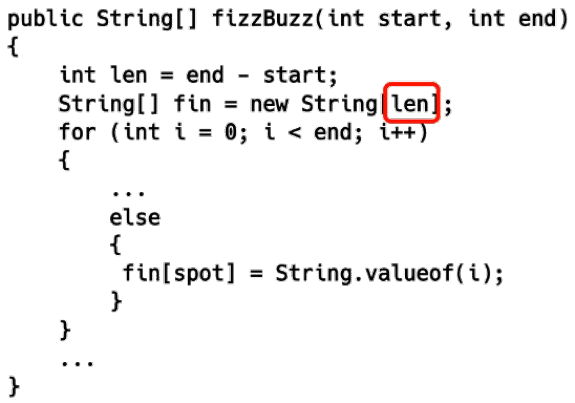} & 
\includegraphics[width=0.22\textwidth]{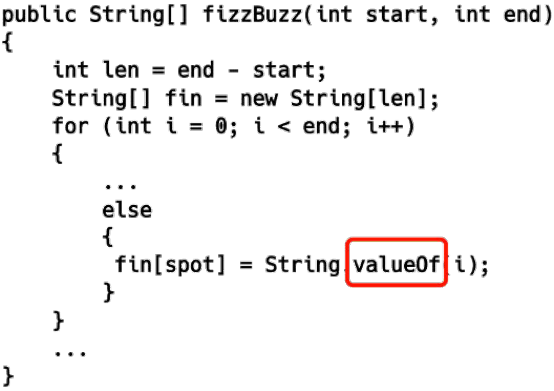} & 
\includegraphics[width=0.22\textwidth]{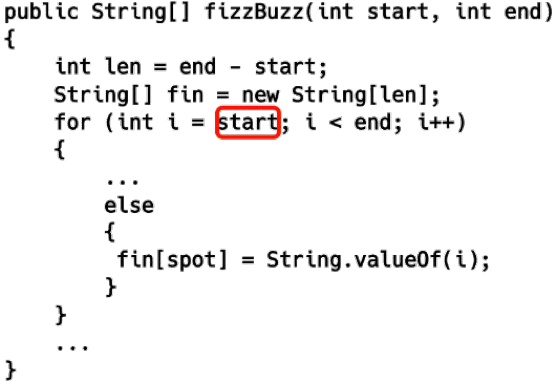} \\
\hline
\end{tabular}
}
\label{table:casestudy}
\vspace{-0.15in}
\end{table*}

\subsection{Experimental Results}
To evaluate the unity of simulation, we designed four tasks to compare different methods. The experimental results are presented in Table \ref{table:result2} and Table \ref{table:result3}, revealing several noteworthy findings. First, our proposed method exhibits a robust ability to adapt to diverse task requirements, as shown in Table \ref{table:result1}. In contrast, traditional KT models, which serve as baselines, are only capable of addressing Task 4, while the OKT model is restricted to Tasks 1, 2, and 3. These limitations underscore the broader applicability and versatility of the CoderAgent framework in handling a wider range of tasks.

For Tasks 1, 2, and 3, CoderAgent consistently outperforms OKT. This superior performance highlights CoderAgent's ability to extract fine-grained insights from students' learning processes. By simulating the iterative nature of code modification, CoderAgent can effectively predict students' next modification intentions and accurately identify specific areas of the code requiring adjustment. These strengths demonstrate the model's capacity to provide a detailed and precise representation of students' problem-solving strategies, offering a significant improvement over prior methods.

In Task 4, which involves predicting whether the next code submission will result in an Accepted outcome, traditional KT models perform poorly, yielding AUC scores near 0.5. This underperformance is largely attributed to the unique characteristics of programming tasks, where students often require multiple attempts to achieve a correct solution. The resulting dataset imbalance, characterized by a significantly higher proportion of incorrect instances compared to correct instances, leads these models to predominantly predict incorrect outcomes. In contrast, CoderAgent achieves notable improvements by leveraging its agent-based framework. Rather than relying on straightforward label fitting, it assesses whether a code submission has the potential to satisfy acceptance criteria. This approach enables more accurate predictions and a balanced performance in Task 4.

Overall, these results demonstrate that CoderAgent outperforms baseline models across all tasks. It provides a nuanced and interpretive understanding of students' learning behaviors, thereby offering significant advantages over traditional methodologies in the domain of programming education.

\subsection{Case Study}
\textbf{Motivation:} To validate the interpretability of CoderAgent in simulating student's code modifications during the iterative process, we choose a question as a case study:
\textit{Consider a series of numbers starting from a given value start and running up to but not including end. The goal is to return a new String[] array containing the string form of these numbers, with specific rules: for multiples of 3, use ``Fizz'' instead of the number; for multiples of 5, use "Buzz"; and for multiples of both 3 and 5, use ``FizzBuzz''. In Java, String.valueOf(xxx) will convert an integer (or other types) to a string.}

We selected a CoderAgent simulation process, summarized in Table \ref{table:casestudy}, omitting code sections that were correct and unchanged. In the student's initial submission, three errors were identified: an undefined variable n, a misspelling of valueOf as valueof, and an incorrect loop starting value. In the first iteration, CoderAgent used compiler feedback to identify and fix the undefined variable. In the second iteration, it corrected the misspelled function, recognizing String.valueOf() as the intended method based on task hints. Once the syntax errors were resolved, CoderAgent focused on the semantic issue within the loop—the starting value was incorrect for the intended sequence. The agent, simulating the student's knowledge level, accurately identified the semantic error in the loop and modified it to produce the correct result.

The entire process is as same as the student's actual modification path: first resolving syntax errors, then addressing semantic issues. This case study highlights CoderAgent's ability to model a student's reasoning in real-time, ensuring both precision in detecting errors and interpretability.

\subsection{Simulations in Programming Applications}
\begin{figure}[t]
  \centering
  \includegraphics[width=0.48\textwidth]{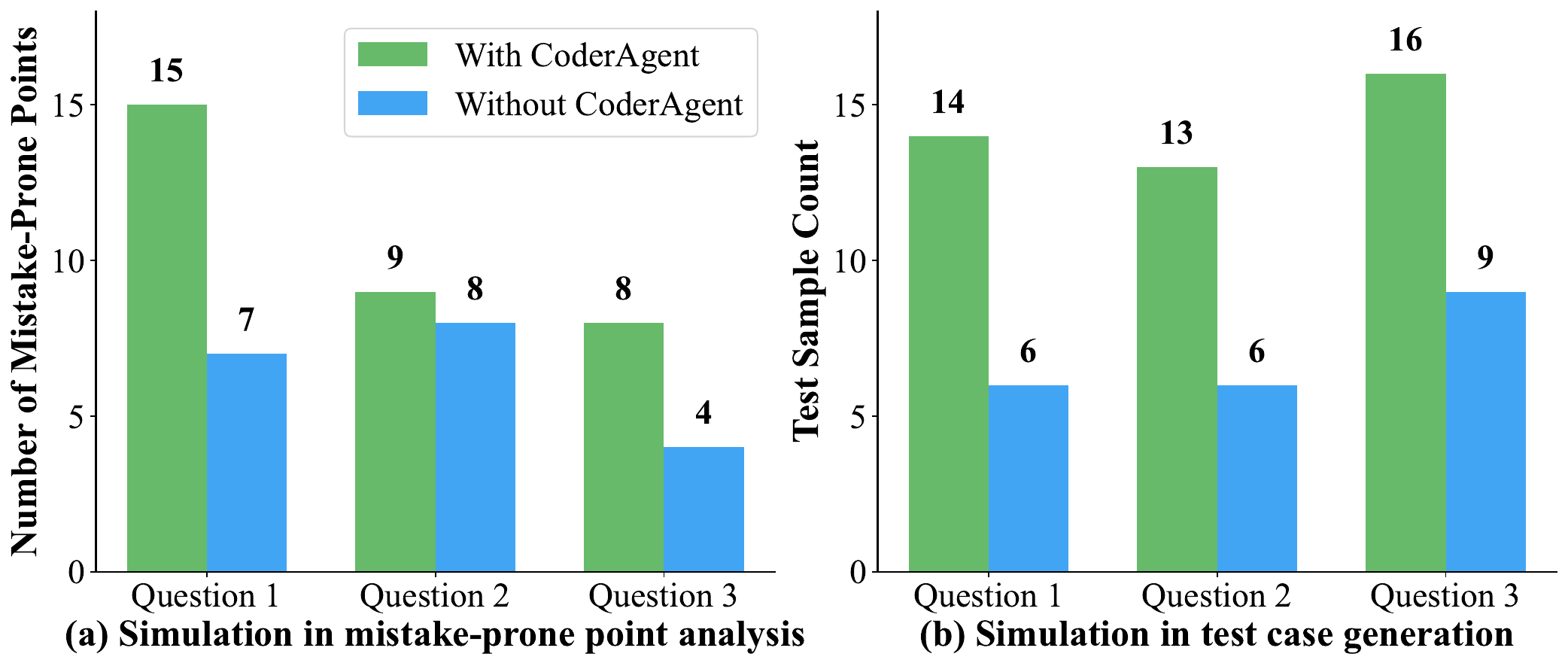}
  \vspace{-0.2in}
  \caption{Simulation results in programming applications}
  \label{fig:simulation}
   \vspace{-0.2in}
\end{figure}
\textbf{Motivation:} Designing experiments that are suitable for students is essential for classroom instruction. To assess the applicability and practicality of CoderAgent, we applied it in simulating a course-based programming experiment.

The experiment involved three programming questions, using data from 20 randomly selected students. For each task, we examined common errors under two conditions: one without CoderAgent and the other with simulated submissions generated by CoderAgent. The results, shown in Figure \ref{fig:simulation} (a), reveal that using simulated data improved the identification of error-prone areas. With CoderAgent, a wider range of common errors was detected, providing deeper insights into the challenges students may encounter. This suggests that the simulated results produced can significantly support the design of course experiments. By identifying a broader spectrum of errors, educators can create assignments that better align with students' learning trajectories, addressing potential difficulties more effectively. Additionally, the increased variety of identified errors enables the development of more robust test cases. Building on this, we conducted a second simulated experiment, where test cases for the problem were generated under both conditions. The experimental results, presented in Figure \ref{fig:simulation} (b), show that CoderAgent generated more comprehensive test cases, ensuring that assignments reinforce students' understanding of key concepts. This highlights the potential of CoderAgent to enhance the quality and effectiveness of programming education.

\section{Conclusion}
In this paper, we presented CoderAgent, a LLM-based agent to simulate students’ iterative coding processes with interpretability and granularity, without relying on extensive datasets. By drawing on the ACT-R theory, we defined programming proficiency as a combination of knowledge and coding skills. Central to our approach is the Programming Tree of Thought, which deconstructs the planning phase into four distinct steps, enabling CoderAgent to effectively model students’ modification intentions and strategies. Experimental evaluations on real-world datasets and programming simulations demonstrated the framework’s effectiveness in capturing and analyzing student coding behaviors.

\section*{Acknowledgements}
This research was supported by grants from the National Key Research and Development Program of China (Grant No. 2024YFC3308200), the National Natural Science Foundation of China (62337001), the Key Technologies R \& D Program of Anhui Province (No. 202423k09020039) and the Fundamental Research Funds for the Central Universities.
\bibliographystyle{named}
\bibliography{ijcai25}

\end{document}